\documentclass[11pt]{article}
\usepackage{naaclhlt2015}
\usepackage{times}
\usepackage{url}
\usepackage{latexsym}
\usepackage{comment}
\usepackage{amsmath,amsthm,amssymb}
\usepackage[vlined]{algorithm2e}
\usepackage{graphicx}
\usepackage{multirow}
\usepackage{color}
\usepackage{tablefootnote}
\usepackage{skt}
\usepackage[utf8]{inputenc}
\usepackage[russian,english]{babel}

\title{Multilingual Open Relation Extraction\\
Using Cross-lingual Projection}

\author{Manaal Faruqui\\
  Carnegie Mellon University \\
  Pittsburgh, PA 15213 \\
  {\tt mfaruqui@cs.cmu.edu} \\\And
  Shankar Kumar \\
  Google Inc. \\
  New York, NY 10011 \\
  {\tt shankarkumar@google.com} \\
}

\begin{document}
\maketitle
\begin{abstract}
  Open domain relation extraction systems identify relation and argument phrases 
  in a sentence without relying on any underlying schema. However, current 
  state-of-the-art relation extraction systems are
  available only for English because of their heavy reliance on 
  linguistic tools such as part-of-speech taggers and dependency parsers. We 
  present a cross-lingual annotation projection method for
  language independent relation extraction. We evaluate our method
  on a manually annotated test set and present results on three
  typologically different languages. We release these manual annotations and
  extracted relations in 61 languages from Wikipedia.
\end{abstract}

\section{Introduction}
\label{sec:intro}

Relation extraction (RE) is the task of assigning a semantic relationship between a pair of arguments. The two major types of RE are closed domain and open
domain RE. While closed-domain RE systems
\cite{Bunescu:2005:SPD:1220575.1220666,Bunescu07learningto,Mintz:2009:DSR:1690219.1690287,P14-1090,berant2014paraphrasing} consider only a closed set of relationships between two arguments,
open domain systems~\cite{Yates:2007:TOI:1614164.1614177,carlson-aaai,ReVerb2011,ollie-emnlp12} use an arbitrary phrase to specify a relationship. In this paper, we focus on open-domain RE for multiple languages. Although there are advantages to closed domain RE~\cite{banko08tradeoffs}, it is expensive to construct a closed set of relation types which would be meaningful across multiple languages.

Open RE systems extract patterns from sentences in a given language to 
identify relations. For learning these patterns, the sentences are analyzed using
a part of speech tagger, a dependency parser and possibly a named-entity recognizer. In languages other than English, these tools are either unavailable or not accurate enough to be used. In comparison, it is easier to obtain
parallel bilingual corpora which can be used to build machine
translation systems~\cite{Resnik:2003:WPC:964751.964753,smith-EtAl:2013:ACL2013}. 

In this paper, we present a system that performs RE on a sentence in a source language by first translating the sentence to English, performing RE in English, and finally projecting the relation phrase back to the source language sentence. Our system assumes the availability of a machine translation system from a source language to English and an open RE system in English but no any other analysis tool in the source language. The main contributions of this work are:
\begin{itemize}
\item A pipeline to develop relation extraction system for any source language.
\item Extracted open relations in 61 languages based on Wikipedia corpus.
\item Manual judgements for the projected relations in three languages.
\end{itemize}

We first describe our methodology for language independent cross-lingual 
projection of extracted relations~(\S\ref{sec:multi}) followed by the 
relation annotation procedure and the results~(\S\ref{sec:expts}).
The manually annotated relations in 3 languages and the automatically 
extracted relations in 61 languages are available at: 
\url{https://www.kaggle.com/shankkumar/multilingualopenrelations15/}.

\section{Multilingual Relation Extraction}
\label{sec:multi}

\begin{figure}[!tb]
  \centering
  \includegraphics[width=\columnwidth]{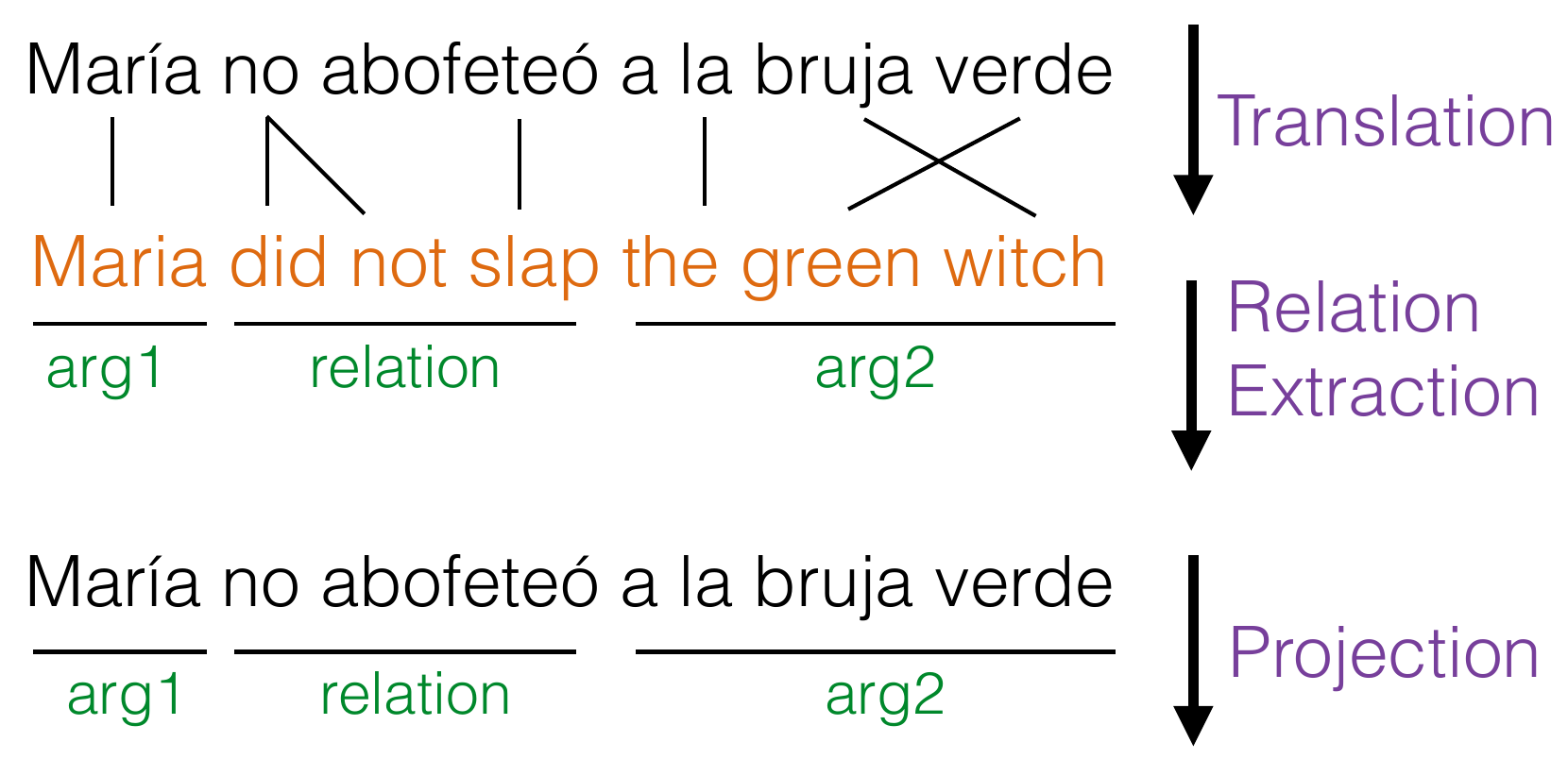}
  \caption{RE in a Spanish sentence using the cross-lingual relation extraction pipeline.}
  \label{fig:example}
\end{figure}

Our method of RE for a sentence $\textbf{s} = \langle s_1, s_2, \dots s_N\rangle$ in a non-English language consists of three
steps: (1) Translation of $\textbf{s}$ into English, that generates a sentence $\textbf{t} = \langle t_1, t_2, \dots t_M\rangle$ with word alignments $\textbf{a}$ relative to $\textbf{s}$, (2) Open RE on $\textbf{t}$, and (3) Relation projection from $\textbf{t}$ to $\textbf{s}$.  Figure~\ref{fig:example} shows an example of RE in Spanish using our proposed pipeline.\footnote{This is a sample sentence and is not taken from Wikipedia.}
 We employ \textsc{Ollie}\footnote{\url{http://knowitall.github.io/ollie/}}~\cite{ollie-emnlp12} for RE in English and \textsc{Google Translate}\footnote{\url{https://developers.google.com/translate/}} API for translation from the source language to English, although in principle, we could use any translation system to translate the language
 to English. 
We next describe each of these components.

\subsection{Relation Extraction in English}
\label{sec:extraction}
Suppose $\textbf{t}=\langle t_1, t_2, \ldots, t_M \rangle$ is a tokenized English sentence. Open relation
extraction computes triples of non-overlapping phrases (\textbf{arg1}; \textbf{rel}; \textbf{arg2}) from the sentence $\textbf{t}$. 
The two arguments $\textbf{arg1}$ and $\textbf{arg2}$ are connected by the relation phrase \textbf{rel}. 

We utilized \textsc{Ollie}~\cite{ollie-emnlp12} to extract the relation tuples for every English sentence. We chose \textsc{Ollie} because it has been shown to give a higher yield at comparable precision relative to other open RE systems such as \textsc{Reverb} and WOE$^{\text{parse}}$~\cite{ollie-emnlp12}.
\textsc{Ollie} was trained by
extracting dependency path patterns on annotated training data. This training data was bootstrapped
from a set of high precision seed tuples extracted from a simpler RE system \textsc{Reverb}~\cite{ReVerb2011}. In \textit{Godse killed Gandhi}, the extracted
relation (Godse; killed; Gandhi) can be expressed by the dependency pattern: \textbf{arg1} $\uparrow$ nsubj $\uparrow$ \textbf{rel}:postag=VBD $\downarrow$ dobj $\downarrow$ \textbf{arg2}.\footnote{Example borrowed from \newcite{ollie-emnlp12}} \textsc{Ollie} also normalizes the relation phrase for some of the phrases, for example \textit{is president of} is normalized to \textit{be president of}. \footnote{For sentences where the veracity of a relation depends on a clause, \textsc{Ollie} also outputs the clause. For example, in \textit{Early astronomers believed that Earth is the center of the universe}, the relation (Earth; be center of; universe) is supplemented by an (\textit{AttributedTo:} believe; Early astronomers) clause. We ignore this clausal information.}

\subsection{Cross-lingual Relation Projection}
\label{sec:projection}

\begin{algorithm}[tb]
 \DontPrintSemicolon
 \KwData{\textbf{s}, \textbf{t}, \textbf{a}, $p_t$}
 \KwResult{$p_s$}
 $P \leftarrow $ PhraseExtract(\textbf{s}, \textbf{t}, \textbf{a})\;
 $p_s = \emptyset$, $score = -\infty$, $overlap=0$\;
 \For{$(phr_s, phr_t) \in P $}{
   \If{$\textrm{BLEU}(phr_t, p_t) > score$}{
      \If{ $phr_t \cap p_t \ne \emptyset$}{
        $p_t \leftarrow phr_t$
        $score \leftarrow \textrm{BLEU}(phr_t, p_t)$
        $overlap \leftarrow phr_t \cap p_t$
      }
   }
 }
 \If{$overlap \ne 0$}{
    $length = \infty$\;
    \For{$(phr_s, p_t) \in P$}{
       \If{$len(phr_s) < length$}{
         $length \leftarrow len(phr_s)$\;
         $p_s \leftarrow phr_s$;
       }
    }
 } \Else {
   $p_s \leftarrow $ WordAlignmentProj(\textbf{s}, \textbf{t}, \textbf{a}, $p_t$);
 }
 \caption{Cross-lingual projection of phrase $p_t$ from a target sentence \textbf{t} to a source sentence \textbf{s}
 using word alignments \textbf{a} and parallel phrases $P$.}
 \label{algo:phrase}
\end{algorithm}
We next describe an algorithm to project the extracted relation tuples in English back to the source language sentence. Given a source sentence, the \textsc{Google Translate} API provides us its translation along with the word-to-word alignments relative to the source. If $\textbf{s} = {s}_{1}^{N}$ and $\textbf{t}={t}_{1}^{M}$ denote the source and its English translation, then the alignment $\textbf{a} = \{a_{ij} : 1 \le i \le N; 1 \le j \le M \}$ where,
$a_{ij}=1$ if $s_i$ is aligned to $t_j$, and is $0$ otherwise. 
A naive word-alignment based projection would map every word from a
phrase extracted in English to the source sentence. This algorithm has
two drawbacks: first, since the word alignments are many-to-many, each
English word can be possibly mapped to more than one source word which leads to ambiguity in its projection; second, a word level mapping can produce non-contiguous phrases in the source sentence, which are hard to interpret semantically.

To tackle these problems, we introduce a novel algorithm that incorporates a BLEU score~\cite{Papineni:2002:BMA:1073083.1073135}  based
phrase similarity metric to perform cross-lingual projection of relations. 
Given a source sentence, its translation, and the word-to-word alignment, we first extract phrase-pairs $P$ using the phrase-extract algorithm~\cite{Och:2004:ATA:1105587.1105589}. In each extracted phrase pair $(phr_s, phr_t) \in P$, $phr_s$ and $phr_t$ are contiguous word sequences in \textbf{s} and \textbf{t} respectively. We next determine the translations of \textbf{arg1}, \textbf{rel} and \textbf{arg2} from the extracted phrase-pairs.

For each English phrase $p \in \{\text{arg1}, \text{rel}, \text{arg2}\}$, we first obtain the phrase-pair $(phr_s, phr_t) \in P$ such that $phr_ t$ has the highest BLEU score relative to $p$ subject to the condition that $p \cap phr_t \ne \emptyset$ i.e, there is at least one word overlap between the two phrases. This condition is necessary since we use BLEU score with smoothing and may obtain a non-zero BLEU score even with zero word overlap. If there are multiple phrase-pairs in $P$ that correspond to the same target phrase $phr_t$, we select the shortest source phrase ($phr_s$). However, if there is no word overlap between the target phrase $p$ and any of the target phrases in $P$, we project the phrase using the word-alignment based projection.  The cross-lingual projection method is presented in Algorithm~\ref{algo:phrase}.

\section{Experiments}
\label{sec:expts}

Evaluation for open relations is a difficult task with
no standard evaluation datasets. We first describe the construction of our 
multilingual relation extraction dataset and then present the experiments.

\paragraph{Annotation.}

The current approach to evaluation for open relations~\cite{ReVerb2011,ollie-emnlp12} is to
extract relations from a sentence and manually annotate
each relation as either valid ($1$) or invalid ($0$) for the sentence. For example, in the sentence: \textit{``Michelle Obama, wife of Barack Obama was born in Chicago''}, the following are possible annotations:
a) (Michelle Obama; born in; Chicago):~$1$, b) (Barack Obama; born in; Chicago):~$0$. 
Such binary annotations are not available for languages apart from English. Furthermore, a binary
$1$/$0$ label is a coarse annotation that could unfairly penalize an extracted relation which has the correct semantics but is slightly ungrammatical. This could occur either when prepositions are dropped from the relation phrase or when there is an ambiguity in the boundary of the relation phrase. 

Therefore to evaluate our multilingual relation extraction framework, we obtained annotations from professional linguists for three typologically
different languages: French, Hindi, and Russian. The annotation task
is as follows: \textit{Given a sentence and a pair of arguments (extracted
automatically from the sentence), the annotator identifies the
most relevant contiguous relation phrase from the sentence that
establishes a plausible connection between the two arguments}.
If there is no meaningful contiguous relation phrase between the two 
arguments, the arguments are considered invalid and hence, the
extracted relation tuple from the sentence is considered incorrect. 

Given the human annotated relation phrase and the automatically
extracted relation phrase, we can measure the similarity between the two,
thus alleviating the problem of coarse annotation in binary judgments.
For evaluation, we first report the percentage of valid arguments. Then
for sentences with valid arguments, we use smoothed sentence-level BLEU 
score (max n-gram order = 3) to measure the similarity of the 
automatically extracted relation relative to the human annotated relation.\footnote{We 
obtained two annotations for $\approx300$ Russian sentences. Between the 
two annotations, the perfect agreement rate was 74.5\% and the average 
BLEU score was $0.85$.}

\begin{table}[t]
  \centering
  \small
  \begin{tabular}{|l||c|c|c|c|}
    \hline
    \multirow{2}{*}{Language} & \multirow{2}{*}{\% valid} & \multirow{2}{*}{BLEU} & 
    \multicolumn{2}{|c|}{Relation length} \\
     & & & Gold & Auto \\
	\hline
	French & 81.6\% & 0.47 & 3.6 & 2.5\\ 
    Hindi & 64.9\% & 0.38 & 4.1 & 2.8 \\ 
    Russian & 63.5\% & 0.62 & 1.8 & 1.7 \\ 
	\hline
   \end{tabular}
   \caption{\% of valid relations and BLEU score of the extracted relations
   across languages with the average relation phrase length (in words).}
  \label{tab:results}
\end{table}

\begin{figure}[t]
  \centering
  \includegraphics[width=0.9\columnwidth]{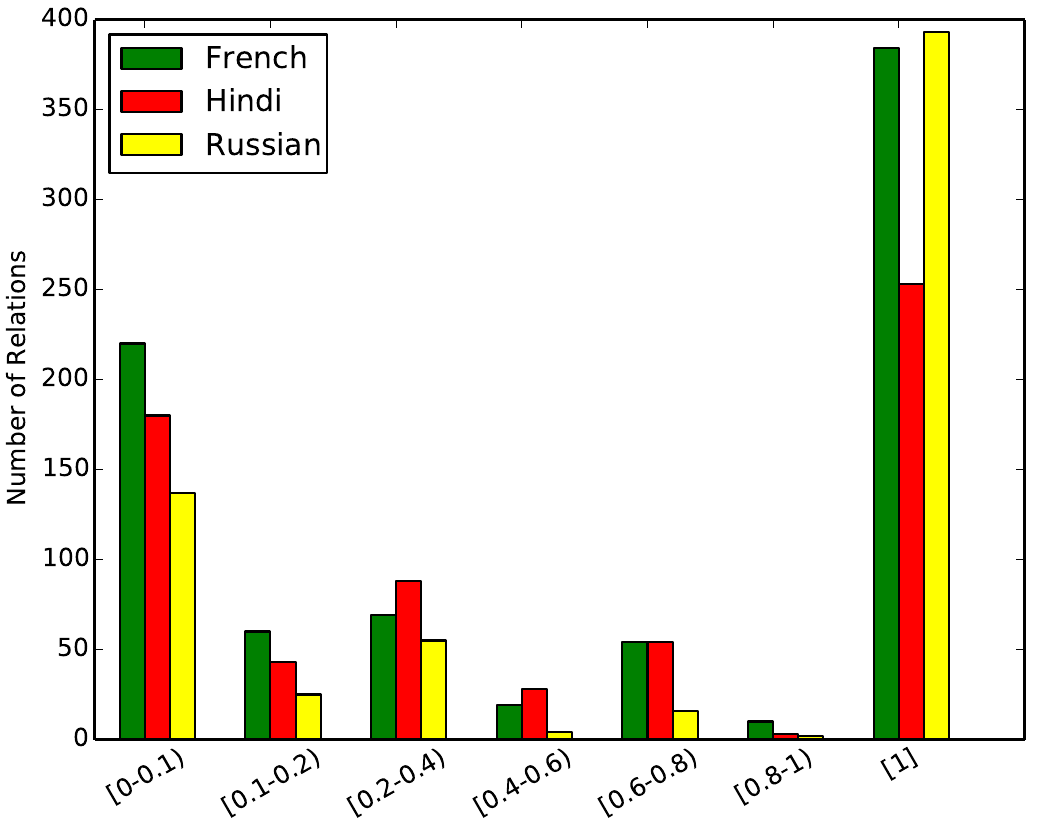}
  \caption{Number of automatically extracted relations binned by their 
  BLEU scores computed relative to the manually annotated relations.}
  \label{fig:bin}
\end{figure}

\paragraph{Results.}
\label{sec:results}

\begin{table}[t]
	\centering
	\small
	\begin{tabular}{|l|r||l|r|}
		\hline
		Language & Size & Language  & Size \\ 
		\hline
		French  & 6,743  & Georgian & 497\\ 
		Hindi  & 367 & Latvian &  491\\ 
		Russian  & 7,532 & Tagalog & 102 \\ 
		Chinese  & 2,876 & Swahili & 114 \\
		Arabic  & 707 & Indonesian & 1,876 \\
		\hline
	\end{tabular}
	\caption{Number of extracted relations (in thousands) from Wikipedia in 
	10 languages out of a total of 61.}
	\label{tab:size}
\end{table}

\begin{table*}[t]
	\centering
	\small
	\begin{tabular}{|l|c|c|c|}
		\hline
		Language & Argument 1 & Relation phrase & Argument 2 \\ 
		\hline
		\multirow{2}{*}{French} & Il & fut enr\^{o}l\'{e} de force au & RAD\\
		& \textit{He} & \textit{was conscripted to} & \textit{RAD}\\
		\hline
		\multirow{2}{*}{Hindi}& bahut se log & aaye & cailifornia \\
		& \textit{Many people} & \textit{came to} & \textit{California}\\
		\hline
		\multirow{2}{*}{Russian} &\begin{otherlanguage*}{russian}Автокатастрофа \end{otherlanguage*} & \begin{otherlanguage*}{russian}произошла\end{otherlanguage*} & \begin{otherlanguage*}{russian}Черногории\end{otherlanguage*}\\
		& \textit{Crash} & \textit{occured} & \textit{Montenegro}\\
		\hline
	\end{tabular}
	\caption{Examples of extracted relations in different languages
		with English translations (Hindi is transliterated).}
	\label{tab:examples}
\end{table*}

We extracted relations from the entire Wikipedia\footnote{\url{www.wikipedia.org}}
corpus in Russian, French and Hindi from all sentences whose lengths are in the range of $10-30$ words. We randomly selected $1,000$ relations for each of these languages and annotated them. The results are shown in table~\ref{tab:results}. The percentage of valid extractions is highest in French ($81.6\%$) followed by Hindi and Russian ($64.0\%$). 
Surprisingly, Russian obtains the lowest percentage of valid relations but has the highest BLEU score between the automatic and the human extracted relations. This could be attributed to the fact that the average relation length (in number of words) is the shortest for Russian. From table~\ref{tab:results}, we observe that the length of the relation phrase is inversely correlated with the BLEU score.

Figure~\ref{fig:bin} shows the distribution of the number of extracted relations across bins of similar BLEU scores. Interestingly, 
the highest BLEU score bin ($1$) contains the maximum number of relations in all three languages. This is an encouraging result since it implies that the majority of the extracted relation 
phrases are identical to the manually annotated relations. Table~\ref{tab:size} lists the
sizes of automatically extracted relations on 61 different languages from Wikipedia
that we are going to make publicly available. These were selected to include
a mixture of high-resource, low-resource, and typologically different languages. Table~\ref{tab:examples} shows examples of randomly selected relations in different languages along with their English translations.

\section{Related Work}
\label{sec:related}

Cross-lingual projection has been used for transfer of syntactic \cite{Yarowsky:2001:IMP:1073336.1073362,Hwa05bootstrappingparsers} and semantic information \cite{Riloff:2002,pado09:annotation}.
There has been a growing interest in RE for languages other than English.
\newcite{Gamallo:2012:DOI:2389961.2389963} present a dependency-parser based open RE system for Spanish, Portuguese and Galician. RE systems for Korean have been developed for both open-domain~\cite{kim-EtAl:2011:IJCNLP-2011} and closed-domain~\cite{kim-lee:2012:ACL2012short,kim:cross-lingual} using annotation projection. These approaches use a Korean-English parallel corpus to project relations extracted in English to Korean. Following projection, a Korean POS-tagger and a dependency parser are employed to learn a RE system for Korean. 

\newcite{E14-4003} describe an open RE for Chinese that employs word segmentation, POS-tagging, dependency parsing.
\newcite{lewis-steedman:2013:EMNLP} learn clusters of semantically equivalent relations across French and English by creating a semantic signature of relations by entity-typing. These relations are extracted using CCG parsing in English and dependency parsing in French. \newcite{Blessing:2012:CDS:2396761.2398411} use inter-wiki links to map relations from a relation database in a pivot language to the target language and use these instances for learning in a distant supervision setting. 
\newcite{gerber2012extracting} describe a multilingual pattern extraction system for RDF predicates that uses pre-existing knowledge bases for different languages.

\section{Conclusion}

We have presented a language independent open domain relation extraction pipeline and have evaluated its performance on three typologically different languages: French, Hindi and Russian. Our cross-lingual projection method utilizes \textsc{Ollie} and \textsc{Google Translate} to extract relations in the language of interest. Our approach does not rely on the availability of linguistic resources such as POS-taggers or dependency parsers in the target language and can thus be extended to multiple languages supported by a machine translation system. We are releasing the manually annotated judgements for open relations in the three languages and the open relations extracted over the entire Wikipedia corpus in 61 languages. The resources are
available at: \url{http://cs.cmu.edu/~mfaruqui/soft.html}.

\section*{Acknowledgment}
This work was performed when the first author was an intern at Google.
We thank Richard Sproat for providing comments on an earlier 
draft of this paper. We thank Hao Zhang for helping us with the relation 
extraction framework, and Richard Zens and Kishore Papineni for their feedback on this
work. We are grateful to Bruno Cartoni, Vitaly Nikolaev and their teams for providing us annotations of multilingual relations.
\bibliography{references}
\bibliographystyle{naaclhlt2015}
\end{document}